%% file: root.tex
\documentclass[letterpaper, 10 pt, conference]{ieeeconf}  % Comment this line out if you need a4paper

\IEEEoverridecommandlockouts                              % This command is only needed if 
\usepackage{booktabs}
\usepackage[table]{xcolor}
\usepackage{vcell}
\usepackage{subcaption}
\usepackage{graphicx}
\usepackage{amsmath}
\usepackage{float}
\usepackage{amssymb}
\usepackage{tikz}
\usepackage{xspace}
\usepackage{bm}
\usepackage{wrapfig}
\usepackage{multirow}
\usepackage{duckuments}
\usepackage{graphicx}
\usepackage{pifont} % for ding
\usepackage{listings}
\usepackage{xcolor} % Required for \textcolor
\usepackage{makecell}
\usepackage{url}

\newcommand{\smallsec}[1]{\noindent {\bf #1.}}
\newcommand{\method}{VLA-0\xspace}
\newcommand{\tb}[1]{\textbf{#1}}
\newcommand{\ul}[1]{\underline{#1}}

\newcommand{\yes}{\color{blue}{\ding{51}}}
\newcommand{\no}{\color{red}{\ding{55}}}
\newcommand{\minus}{\scalebox{0.75}[1.0]{$-$}}
\newcommand{\plus}{\scalebox{0.75}[1.0]{$+$}}
\newcommand{\tbx}[1]{\textcolor{black!40}{#1}}

\title{\LARGE \bf
\method: Building State-of-the-Art VLAs with Zero Modification
}

\author{
Ankit Goyal, Hugo Hadfield, Xuning Yang, Valts Blukis, Fabio Ramos \\
NVIDIA
}

\begin{document}
\maketitle
\thispagestyle{empty}
\pagestyle{empty}

\input{sec/0_abstract}    
\input{sec/1_intro}
\input{sec/2_related_work}
\input{sec/3_method}
\input{sec/4_experiments}
\input{sec/5_conclusions}
\bibliographystyle{plain} \bibliography{main}

%%%%%%%%%%%%%%%%%%%%%%%%%%%%%%%%%%%%%%%%%%%%%%%%%%%%%%%%%%%%%%%%%%%%%%%%%%%%%%%%

\end{document}

%% file: sec/0_abstract.tex
\begin{abstract}
Vision-Language-Action models (VLAs) hold immense promise for enabling generalist robot manipulation. However, the best way to build them remains an open question. Current approaches often add complexity, such as modifying the existing vocabulary of a Vision-Language Model (VLM) with action tokens or introducing special action heads. Curiously, the simplest strategy of representing actions directly as text has remained largely unexplored. This work introduces \method to investigate this idea. We find that \method is not only effective; it is surprisingly powerful. With the right design, \method outperforms more involved models. On LIBERO, a popular benchmark for evaluating VLAs, \method outperforms all existing methods trained on the same robotic data, including $\pi_0.5$-KI, OpenVLA-OFT and SmolVLA. Furthermore, without large-scale robotics-specific training, it outperforms methods trained on large-scale robotic data, like $\pi_0.5$-KI, $\pi_0$, GR00T-N1 and MolmoAct. These findings also translate to the real world, where \method outperforms SmolVLA, a VLA model pre-trained on large-scale real data. This paper summarizes our unexpected findings and spells out the specific techniques required to unlock the high performance of this simple yet potent VLA design. 
% Visual results are provided via the supplementary material.
Visual results, code, and trained models are provided at: https://vla0.github.io/.
\end{abstract}

%% file: sec/1_intro.tex
\section{Introduction}
\label{sec:intro}
\begin{figure*}[tb]
  \centering
  \makebox[\linewidth]{\includegraphics[width=\linewidth]{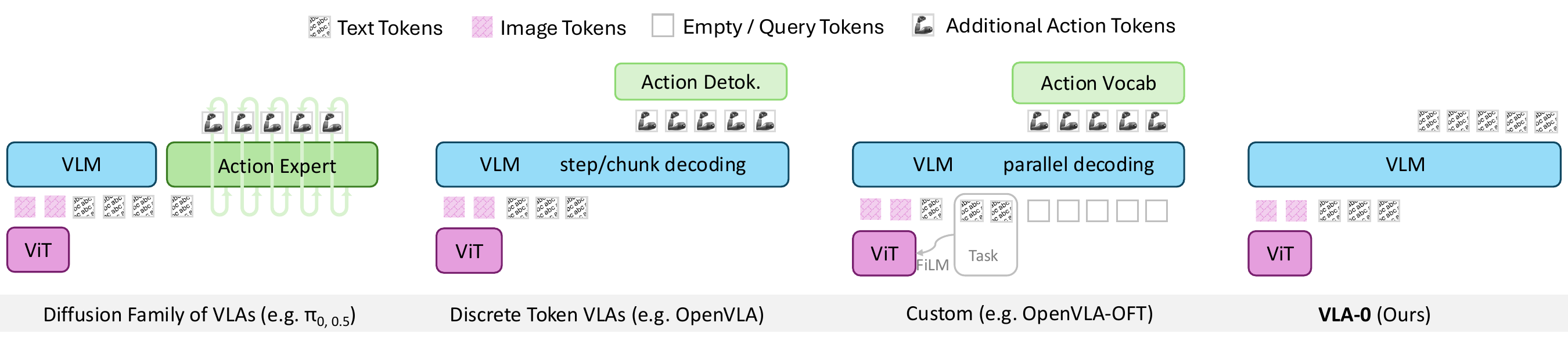}}
   \caption{\tb{Families of methods for building VLAs.} We categorize existing VLAs into three categories: Discrete Token VLAs, Generative Action Head VLAs and Custom Architecture VLAs. In this work, we propose the \method family where the VLM is prompted to directly predict action as text. Unlike other methods, \method requires no change to the underlying VLM.}
   \label{fig:comparison}
\end{figure*}

Following the success of Large Language Models (LLMs) in text processing and Vision-Language Models (VLMs) in handling both visual and textual inputs, a natural next step is to explore Vision-Language-Action models (VLAs), i.e. systems that not only understand visual and textual information, but also predict actions for robotic agents. VLAs are typically built by modifying a base VLM to predict actions. However, it is still unclear what the `correct' way to do this is, if there is one at all. Recent research has taken various approaches, which we broadly categorize into three families, as shown in Figure~\ref{fig:comparison}: (1) Discrete Token VLAs, (2) Generative Action Head VLAs, and (3) Custom Architecture VLAs.

\smallsec{Discrete Token VLAs}
It is one of the initial strategies popularized by models such as RT-2~\cite{zitkovich2023rt} and OpenVLA~\cite{kim2024openvla}. Robot actions, originally continuous, are discretized into bins; each bin is then assigned a token from the VLM vocabulary, using either new or infrequent tokens. The model is then trained to predict these action tokens using the same cross-entropy loss as used to train the base VLM. Although straightforward, this approach has two main limitations: (i) it restricts the resolution of the action space, since fine-grained control can require thousands of bins, which conflicts with sharing the text vocabulary; and (ii) it compromises the pretrained language understanding of the VLM by repurposing its vocabulary for actions. Given these limitations, such VLAs do not perform as well as other alternatives. (see Tab.~\ref{table:main})

\smallsec{Generative Action Head VLAs}
Another common strategy is to attach an action generation head on top of the VLM, as done by methods like $\pi_0$~\cite{black2410pi0} or SmolVLA~\cite{shukor2025smolvla}. The VLM is fine-tuned to predict a latent vector, which is then decoded into actions using a generative model such as a diffusion process or flow matching. While this method improves action fidelity, it also introduces a new neural network that needs to be finetuned.
This often leads to a decline in the language understanding and grounding capabilities of the underlying VLM~\cite{intelligence2025pi_}, and introducing a non-pretrained action head may compromise generalization of the overall system.

\begin{figure}[b!]
  \centering
  \makebox[\linewidth]{\includegraphics[width=0.85\linewidth]{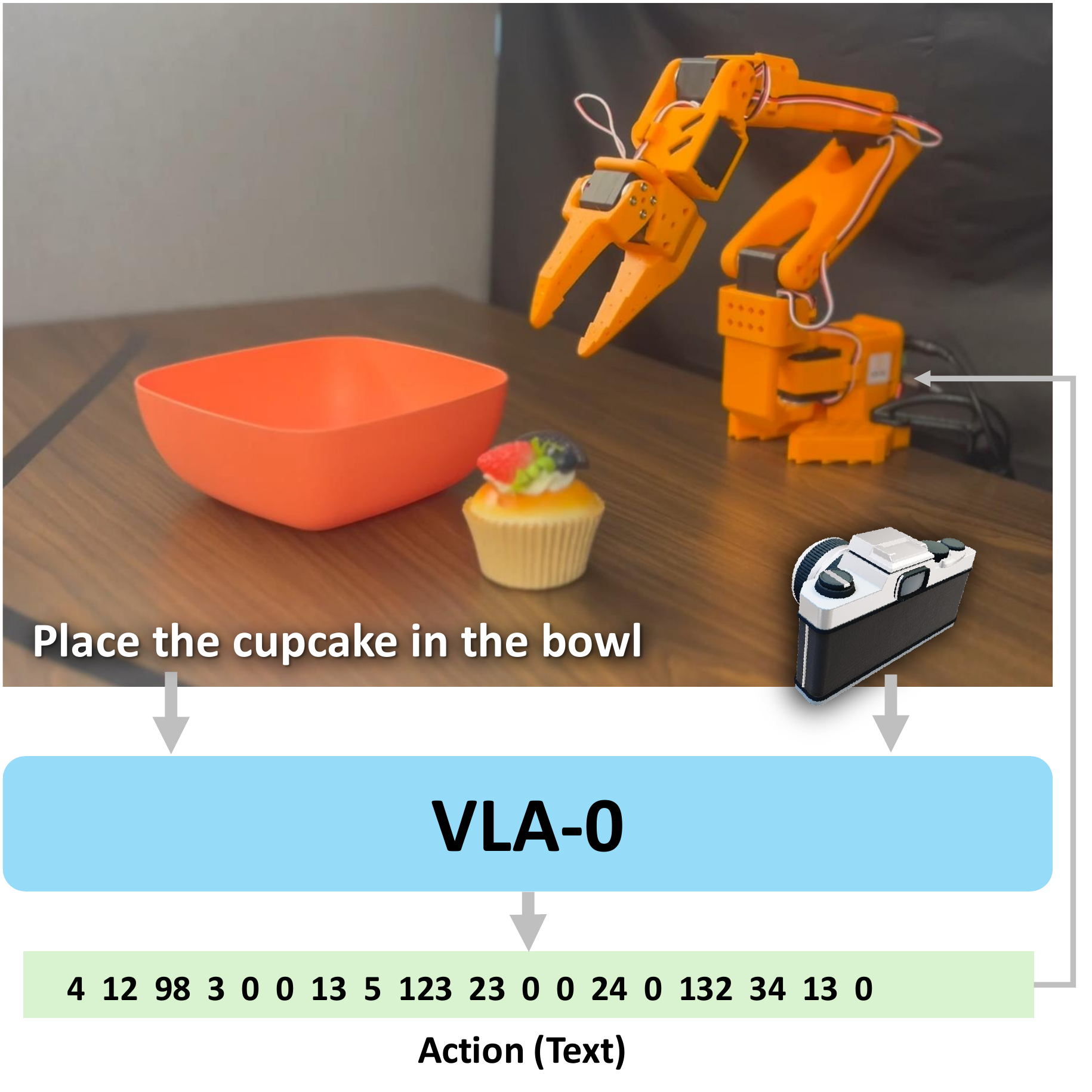}}
   \caption{\textbf{Schematic representation of \method.} \method converts a VLM into a VLA by prompting the VLM to predict action as text. This strategy is surprisingly effective and achieves state-of-the-art results akin to alternatives.}
   \label{fig:teaser}
\end{figure}

\smallsec{Custom Architecture VLAs}
Beyond the above categories, some methods propose architectural modifications or custom tokenizers tailored to action prediction. For instance, OpenVLA-OFT~\cite{kim2025fine} introduces a specialized ACT head. Another example is $\pi$-FAST~\cite{pertsch2025fast} that create a special tokenization scheme for actions using discrete cosine transform (DCT). $\pi$-FAST can also be considered a discrete token VLA, but for the purposes of this work, we classify them as custom VLA as it involves a custom tokenization scheme. While these custom methods are effective, they typically involve significant architectural changes, additional parameters, or custom training pipelines. 

Despite the success of these methods, we ask if there is a simpler alternative. One that does not require changing the VLM's vocabulary or introducing any new architectural components. Have we ruled out predicting actions as text? Why not represent actions (e.g., coordinates, joint angles) as numerical strings and generate them using the VLM’s native text generation capability? This approach does not require new tokens, no vocabulary modifications, and no architectural changes. It maintains the integrity of the VLM while offering arbitrary resolution in the action space. Given the extensive effort devoted to optimizing training recipes for various VLA designs, one may ask: what if we instead focus on the simplest architecture?

We evaluate such a design, which we refer to as \method. Contrary to expectations from prior literature, we find that this simple formulation is highly competitive--achieving performance on par with other alternatives, while requiring no change to the underlying VLM architecture. On the widely adopted LIBERO~\cite{liu2023libero} benchmark, it outperforms all methods that have been trained with the same amount of robotic action data. Further, \method outperforms popular methods that have been pretrained with large-scale action data including $\pi$-0~\cite{black2410pi0}, GR00T-N1~\cite{bjorck2025gr00t},
$\pi$-Fast~\cite{intelligence2025pi_},  OpenVLA~\cite{kim2024openvla},  Octo~\cite{team2024octo} and MolmoAct~\cite{lee2025molmoact}. We find that these findings also translates to the real-world where \method outperforms SmolVLA~\cite{shukor2025smolvla} which has previously achieved state-of-the-art results.

In order to achieve state-of-the-art performance with this design, a careful training and testing recipe is required. For example, we find that during training, random masking of the action text improves performance. Similarly, during testing, it is helpful to ensemble previous predictions. 
\noindent
In summary, our contributions are as follows: \medskip\\
{1:} We demonstrate that a simple VLA design that requires no change to the VLM architecture can achieve state-of-the-art results akin to popular alternatives;\\
{2:} We devise the training and testing recipe that achieves state-of-the-art performance with the simple VLA design.

%% file: sec/2_related_work.tex
\section{Related Work}
\label{sec:related}
\begin{figure*}[h!t!]
  \centering
   \includegraphics[width=\linewidth]{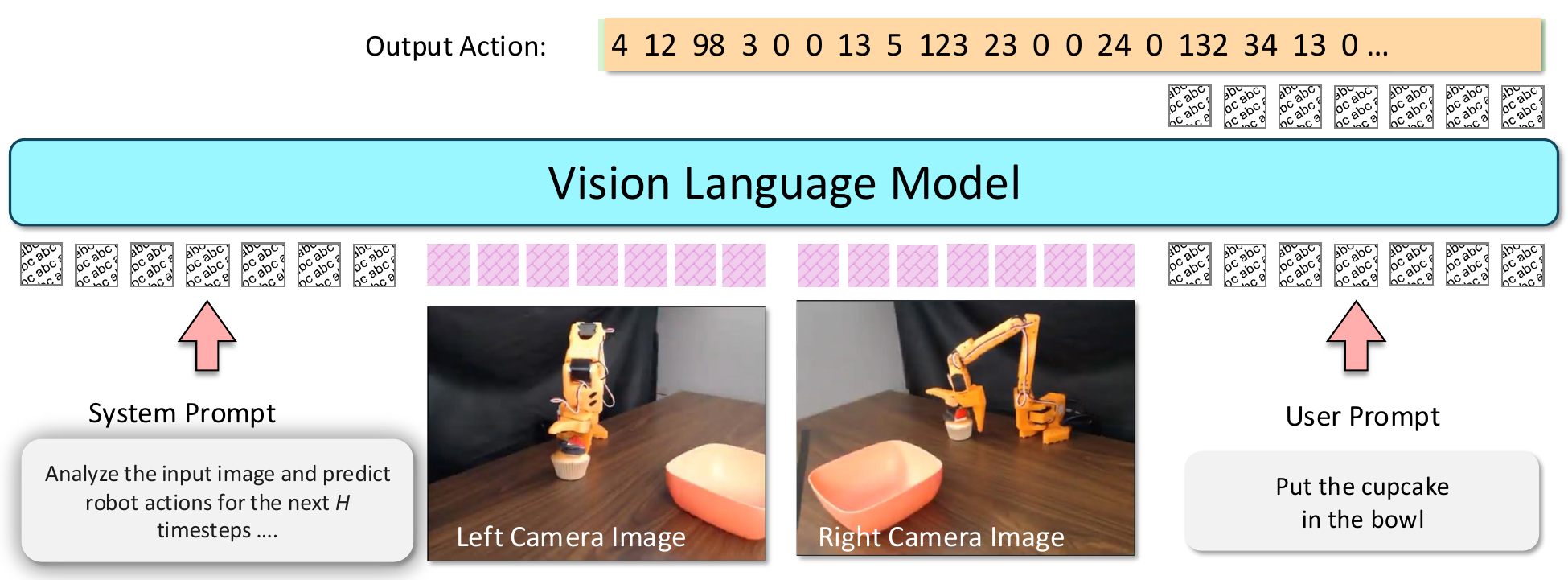}
   \caption{\textbf{Our proposed \method}. It creates a VLA without making any changes to the underlying VLM. It takes a system prompt, language instruction, and images as input, and outputs actions represented as space-separated integers.}
   \label{fig:onecol}
\end{figure*}
Our work builds on recent advances in Vision-Language-Action models and the broader field of robot learning.

\smallsec{Vision-Language-Action Models}
The paradigm of adapting pre-trained Vision-Language Models (VLMs) for robotic control has recently gained significant traction. A predominant approach involves representing continuous actions as discrete tokens. This strategy, employed by influential models like RT-2~\cite{zitkovich2023rt} and OpenVLA~\cite{kim2024openvla}, discretizes the action space into a finite number of bins and maps each bin to a token within the VLM's vocabulary. While this allows for a straightforward integration of action generation into the language modeling objective, it introduces a trade-off between action resolution and vocabulary size, and can potentially corrupt the semantic meaning of repurposed tokens.

Another prominent family of methods avoids altering the VLM's vocabulary by introducing auxiliary action heads. Models such as $\pi_0$~\cite{black2410pi0} and SmolVLA~\cite{shukor2025smolvla} finetune the VLM to output a latent embedding, which is then decoded into a continuous action by a separate generative model, like a diffusion policy or a flow-matching network. While this preserves the VLM's original vocabulary and allows for high-fidelity actions, it increases model complexity and can sometimes lead to a degradation of the VLM's language grounding capabilities~\cite{intelligence2025pi_}. 

A third category involves more substantial architectural modifications [\cite{kim2025fine,pertsch2025fast,singh2025og}], such as the specialized action head in OpenVLA-OFT~\cite{kim2025fine} or the custom action tokenization via Discrete Cosine Transform in $\pi$-FAST~\cite{pertsch2025fast}, which often require intricate training pipelines.

Our proposed method, \method, explores a conceptually simpler alternative: representing actions directly as text. By representing numerical actions (e.g., end-effector coordinates) as strings, we leverage the VLM's native text generation capabilities without any architectural modifications. The closest to our method is LLARVA~\cite{niu2024llarva} that learns to predict action as text. However, LLARVA employs a two-stage process, first generating a 2D trajectory plan before predicting the final action. In contrast, our work demonstrates that direct, end-to-end generation of action strings can achieve state-of-the-art performance. The key to our success lies in a carefully designed training and inference recipe, including action token masking and prediction ensembling, a critical component not explored in LLARVA.

Another close work is HAMSTER~\cite{li2025hamster}, which proposes a hierarchical vision-language-action model. The first stage of HAMSTER using a VLM to predict a 2D action trajectory in text. Our design is similar but we predict the complete robot action (like joint pose or end-effector delta) as text.

\smallsec{Robot Learning Policies}
Learning robotic policies from demonstrations is a well-established field that predates the recent rise of VLAs. Unlike VLAs, these methods typically train policies from scratch on in-domain data without leveraging large pretrained vision and language models. A leading example is Diffusion Policy~\cite{chi2023diffusion}, which models the action space using a conditional diffusion process and has shown strong performance across various manipulation tasks.

Another line of research focuses on improving sample efficiency and spatial reasoning by incorporating explicit 3D representations into the policy architecture [\cite{goyal2023rvt,goyal2024rvt,qian20243d,gervet2023act3d}]. Models like RVT~\cite{goyal2023rvt}, RVT-2~\cite{goyal2024rvt}, ManiFlow~\cite{yan2025maniflow} and Act3D~\cite{gervet2023act3d} leverage 3D scene information to learn more robust and generalizable policies.

In contrast to these approaches, \method aligns with the VLA paradigm by building directly on the powerful, pre-trained representations of a VLM. Our findings indicate that by properly harnessing the VLM, our simple approach can outperform specialized methods like Diffusion Policy on benchmark tasks using only the in-domain action data.

%% file: sec/3_method.tex
\section{Method}
\label{sec:method}
\subsection{Background}
\smallsec{Vision-Language Models}
Vision-Language Models (VLMs) are a class of neural networks designed to process and reason about information from both visual and textual modalities. Typically, they are comprised of a pretrained vision encoder (e.g., a Vision Transformer) that extracts visual features from an image, and a Large Language Model (LLM) that processes textual information. The visual features are projected into the LLM's embedding space, allowing the model to jointly condition on both the image and a text prompt to generate a coherent textual output. \footnote{Some VLMs can also generate outputs in other modalities, such as images. However, for clarity in this paper, we refer to VLMs as models that produce only text.}

For this work, we build our system upon a publicly available, state-of-the-art VLM. Specifically, we employ the 3-billion-parameter Qwen-VL-2.5~\cite{qwen2024qwen2} model, although our method is applicable to any other VLM. Several factors motivate our choice. Qwen-VL-2.5-3B demonstrates highly competitive performance for its model size. As a relatively smaller VLM, it is computationally efficient, which facilitates faster training and inference. Furthermore, its open-weight nature promotes accessibility and reproducibility.

\subsection{Method: \method}
We introduce \method, a simple design for building Vision-Language-Action models. Unlike alternatives, \method preserves the integrity of the underlying VLM: it does not introduce new tokens, alter the existing vocabulary, or add any new neural network layers. Despite its simplicity, and contrary to expectations from prior literature, \method is as performant as more involved alternatives. However, achieving this performance relies on a careful recipe. Three key components of this recipe are action decoding, ensemble prediction, and masked action augmentation.
\input{tables/main}

\smallsec{Input}
\method inherits the input structure of the underlying VLM, which consists of a \textit{System Prompt}, \textit{Images}, and a \textit{Task Instruction}. The system prompt specifies the high-level goal of the VLM. During fine-tuning, we use the following prompt, where $H$, $D$, and $B$ are chosen based on the data.

\textbf{System Prompt.} \textit{Analyze the input image and predict robot actions for the next $H$ timesteps. Each action has $D$ dimensions. Output a single sequence of $H \times D$ integers (0 - $B$ each), representing the $H$ timesteps sequentially. Provide only space-separated numbers. Nothing else.}

Similar to the underlying VLM, \method can take one or more images as input, depending on the setup. For our simulation experiments, we use the third person and wrist camera image as input, like the baselines. For real experiments, we use the left and right camera images as shown in Figure.~\ref{fig:onecol}. We also experiment with an alternative image input design where, instead of providing images as separate entities, we tile them into a single composite image. In our experiments, we find that both designs exhibit similar performance (see Table~\ref{table:ablation}). Lastly, the input includes the task instruction, for example: ``put the banana on the plate."
\smallsec{Action Decoding}
\method produces actions as text. To simplify this task, we ask the VLM to output actions as integers. Specifically, the original continuous action values are first normalized to a fixed integer range (e.g., [0,1000]). The VLM is then tasked with generating an integer for each action dimension. The maximum value of this range can be tuned based on the dataset and the desired action resolution. Notably, unlike discrete token-based VLAs, this approach allows for arbitrary resolution without altering the model's vocabulary.

\smallsec{Ensemble Prediction}
\method employs the prediction ensembling technique introduced by the Action-Chunking Transformer (ACT)~\cite{zhao2023learning}, which has also been adopted by other state-of-the-art VLAs like OpenVLA-OFT~\cite{kim2025fine}. At each inference step, the VLM predicts a sequence of $n$ future actions. Consequently, for the current time step $t$, there are $n$ available predictions for the action: one made at the current step $t$, another made at step $t \minus 1$ (as the second action in its predicted sequence), and so on, back to the prediction made at step $t \minus n \plus 1$. In our design, we average these $n$ predictions to produce the final, more stable action at time step $t$.

\smallsec{Masked Action Augmentation}
Another component of our recipe is a training augmentation we introduce, which we call Masked Action Augmentation. VLMs produce text auto-regressively, meaning each generated token is conditioned on the previously generated tokens. During training, we randomly mask out characters in the target action string. This procedure forces the VLM to reason about the action based on the visual observation and instruction, rather than simply relying on auto-completing a numerical sequence it has started to generate.

\smallsec{Training Details}
We train \method by performing a full fine-tuning of the base VLM. The model is trained to generate the target action strings using a standard cross-entropy loss over the vocabulary. For optimization, we use the Adam optimizer and train the model for 64 epochs with a batch size of 192 and a learning rate of 5e-6. Training takes approximately 32 hours on 8 A100 GPUs.

%% file: tables/main.tex
\begin{table*}[t]
\setlength{\abovetopsep}{5pt}
\centering
\begin{tabular}{l c c c c c c c c}
\toprule
Models                                                                 & \thead{Large-scale \\ act. pre-train} & \thead{VLA \\ Type} & \thead{Spatial} & \thead{Object} & \thead{Goal}      & \thead{Long}      & \thead{Avg.}      & \thead{Avg. \\ rank} \\
\midrule
Diffusion Policy~\cite{chi2023diffusion,kim2024openvla}                 & \no                                   & N/A                 & 78.3            & 92.5           & 68.3              & 50.5              & 72.4              & 6.5 \\
$\pi_{0}$-FAST (Paligemma)~\cite{black2410pi0,shukor2025smolvla}        & \no                                   & Custom              & 87.0            & 63.0           & 89.0              & 48.0              & 71.8              & 6.0 \\
SmolVLA (0.24B)~\cite{shukor2025smolvla}                                 & \no                                   & Gen Head            & 87.0            & 93.0           & 88.0              & 63.0              & 82.8              & 5.3 \\
SmolVLA (2.25B)~\cite{shukor2025smolvla}                                 & \no                                   & Gen Head            & 93.0            & 94.0           & 91.0              & 77.0              & 88.8              & 4.0 \\
OpenVLA-OFT~\cite{kim2025fine}                                         & \no                                   & Custom              & 94.3            & 95.2           & 91.7              & \ul{86.5}         & 91.9              & 2.8 \\
$\pi_{0.5}-KI$~\cite{driess2025knowledge}                               & \no                                   & Gen Head            & \ul{96.6}       & \ul{97.2}      & \ul{94.6}         & 85.8              & \ul{93.3}         & \ul{2.3} \\
\method (Ours)                                                         & \no                                   & Simple              & \tb{97.0}       & \tb{97.8}      & \tb{96.2}         & \tb{87.6}         & \tb{94.7}         & \tb{1.0} \\
\midrule
Octo~\cite{team2024octo}                                               & \yes                                  & Gen Head            & 78.9            & 85.7           & 84.6              & 51.1              & 75.1              & 8.8 \\
OpenVLA~\cite{kim2024openvla}                                          & \yes                                  & Dis. Tok.           & 84.7            & 88.4           & 79.2              & 53.7              & 76.5              & 8.0 \\
$\pi_{0}$-FAST~\cite{pertsch2025fast}                                  & \yes                                  & Custom              & 90.0            & 86.0           & 95.0              & 73.0              & 86.0              & 6.5 \\
Molmo Act~\cite{lee2025molmoact}                                       & \yes                                  & Dis. Tok.           & 87.0            & 95.4           & 87.6              & 77.2              & 86.8              & 6.5 \\
GR00T-N1~\cite{bjorck2025gr00t}                                         & \yes                                  & Gen Head            & 94.4            & 97.6           & 93.0              & \ul{90.6}         & 93.9              & 4.5 \\
$\pi_{0}$~\cite{black2410pi0}                                          & \yes                                  & Gen Head            & 96.8            & \tb{98.8}      & 95.8              & 85.2              & 94.2              & 3.3 \\
$\pi_{0.5}-KI$~\cite{driess2025knowledge}                               & \yes                                  & Gen Head            & \tb{98.0}       & 97.8           & 95.6              & 85.8              & 94.3              & 3.0 \\
OpenVLA-OFT~\cite{kim2025fine}                                         & \yes                                  & Custom              & \ul{97.6}       & \ul{98.4}      & \tb{97.9}         & \tb{94.5}         & \tb{97.1}         & \tb{1.5} \\
\tbx{\method (Ours)}                                                   & \tbx{\no}                             & \tbx{Simple}        & \tbx{97.0}      & \tbx{97.8}     & \tbx{\ul{96.2}}   & \tbx{87.6}        & \tbx{\ul{94.7}}   & \tbx{\ul{2.8}} \\
\bottomrule
\end{tabular}
\caption{\textbf{Performance on Libero.} We report the success rate for four LIBERO suites, for two category of models, with and without large scale action pretraining. Best performance is highlighted in \tb{bold} and second best in \ul{underline}. \method outperforms all models without large scale action pretraining, achieving the highest average success rate and best rank. Further, \method, without any large-scale action pretraining, outperforms popular VLAs including $\pi_{0.5}-KI$, $\pi_0$ and GR00T-N1 that have been pretrained with large scale action data.}
\label{table:main}
\end{table*}

%% file: sec/4_experiments.tex
\section{Experiments}
\label{sec:exp}
\subsection{Setup}
We evaluate our model in both real-world and simulated environments to thoroughly assess its performance.
\begin{figure*}[!t]
  \centering
  \vspace{1mm}
   \includegraphics[width=\linewidth]{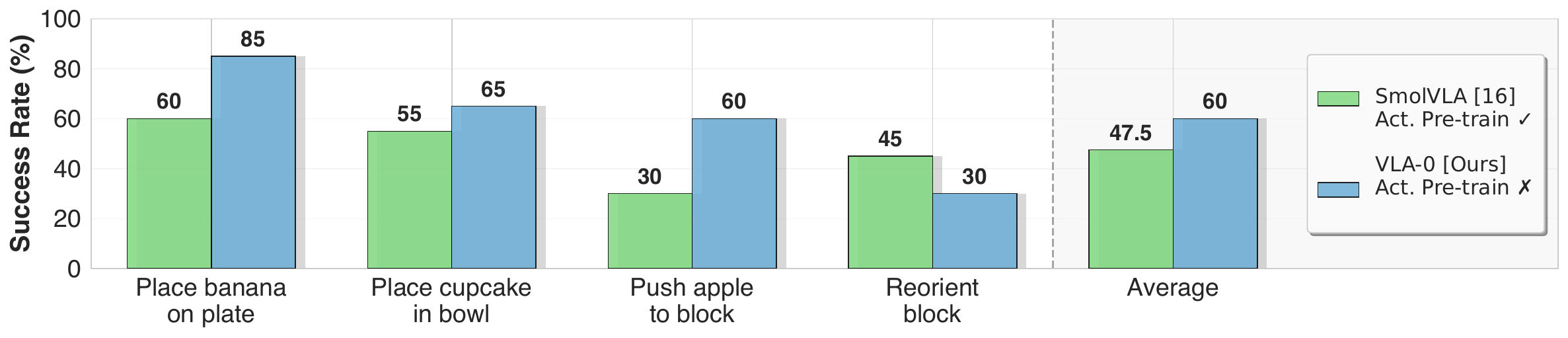}
   \caption{\textbf{Performance on Real.} We compare \method to SmolVLA on four different tasks with SO-100. \method outpeforms SmolVLA on average. SmolVLA is pretrained with large-scale SO-100 data while \method is trained from scratch.}
   \label{fig:real}
\end{figure*}
\smallsec{Real-World}
For our real-world evaluation, we use the SO-100 robot and the LeRobot framework. We train and test policies on four distinct manipulation tasks: reorienting a block, pushing an apple, picking and placing a banana, and picking and placing a cupcake. For each task, we collect 100 demonstrations for training. The learned policies are then evaluated across varied initial conditions of the objects to test for robustness.

\smallsec{Simulation}
In simulation, we use the LIBERO benchmark~\cite{liu2023libero}, a widely adopted benchmark for comparing VLA models. LIBERO consists of four suites: Spatial, Object, Goal, and Long. Each suite is designed to assess a system's capability along a particular dimension. Each suite contains 10 tasks, and each task is tested over 50 episodes. Performance is reported as the success rate for each individual suite and as an overall average.

\subsection{Baselines}
We compare \method against several baselines, including Diffusion Policy and a variety of state-of-the-art Vision-Language-Action (VLA) models. Our baselines are drawn from the three categories previously outlined (Sec.~\ref{sec:intro}, Figure~\ref{fig:comparison}): OpenVLA~\cite{kim2024openvla} and MolmoAct~\cite{lee2025molmoact} from the discrete token-based family; Octo~\cite{team2024octo}, $\pi_0$~\cite{black2410pi0}, GR00T-N1~\cite{bjorck2025gr00t}, $\pi_{0.5}-KI$~\cite{intelligence2025pi_} and SmolVLA~\cite{shukor2025smolvla} from the generative action-based family; and $\pi_0$-FAST~\cite{pertsch2025fast} and OpenVLA-OFT~\cite{kim2025fine} from the custom architecture family. By $\pi_{0.5}-KI$, we refer to the work by Dries et al. which builds on top of $\pi_{0.5}$ to show the effectiveness of knowledge insulation in VLAs. A key distinction among these VLA models is their use of large-scale action pretraining. To ensure a fair comparison, our primary analysis focuses on models that, like \method, have not undergone such pretraining. However, for completeness, we also report the performance of pretrained models in Table~\ref{table:main}.

\subsection{Simulation Results}
Table.~\ref{table:main} summarizes the performance of various baselines and \method on LIBERO. We find the \method outperforms all existing VLA models that, like our, were not pretrained with large-scale robotic data. These include models like $\pi_{0.5}$-KI, OpenVLA-OFT~\cite{kim2024openvla}, $\pi_0$-Fast~\cite{pertsch2025fast} and SmolVLA~\cite{shukor2025smolvla}. \method outperforms these baseline across all the LIBERO suites, outperforming the second best method by 1.4 points on average. This result is highly surprising and runs counter to the expectations set by existing literature. It shows how highly performance VLAs can be built without introducing any change to the underlying VLM.

What's even more surprising is how \method stacks up against models that did have the advantage of pretraining on large-scale robotic data. Despite having no large-scale action pretraining, \method surpasses the performance of many well-known pretrained models, including $\pi_{0.5}$-KI, $\pi_0$, $\pi_0$-FAST~\cite{pertsch2025fast}, Octo~\cite{team2024octo}, OpenVLA~\cite{kim2024openvla} and GROOT-N1.5~\cite{bjorck2025gr00t} and MolmoAct~\cite{lee2025molmoact}. Overall, it gets the second best average rank 2.8, trailing only OpenVLA-OFT~\cite{kim2025fine} (average rank 1.5), a custom VLA model. This suggests that proposed simple strategy holds up effectively against the best models that are pretrained with large scale training data.

\subsection{Real-World Evaluation}
To validate our approach on physical hardware, we evaluate \method in the real world using the LeRobot framework~\cite{cadene2024lerobot}. We compare with SmolVLA~\cite{shukor2025smolvla}, a strong baseline that was specifically trained on the large-scale SO-100 dataset and has been shown to outperform popular methods like $\pi_0$~\cite{black2410pi0} and ACT~\cite{zhao2023learning} on this platform.

For inference, we use a desktop equipped with a 5090 GPU. Our system streams actions for each timestep, achieving an inference speed of 4 Hz. This performance is achieved using a standard PyTorch implementation. We believe this speed could be significantly increased through techniques such as model distillation or quantization, which we leave as future work. For simplicity, we do not ensemble actions in real, although it is possible to do so but requires 8 simultaneous running instances of the model.

Figure~\ref{fig:real} summarizes the task success rates on four real world tasks. The results show that \method outperforms SmolVLA by 12.5 points, despite not being pretrained on the large-scale SO100 dataset. This demonstrates that our method's effectiveness translates from simulation to real.

\subsection{Ablations}
We conduct a series of ablation studies on the LIBERO benchmark to analyze the impact of key design components in \method. Table~\ref{table:ablation} summarizes these results.

\paragraph{Action Ensembling}
Disabling action ensembling (comparing Row 0 and 1) reveals its significant impact. We find that this technique is a critical component, improving the overall success rate by 2 points.

\paragraph{Masked Action Augmentation}
Our proposed Masked Action Augmentation provides a modest but consistent benefit. Removing this augmentation (comparing Row 0 and 2) decreases the success rate by 1.2 points.

\paragraph{Action Resolution}
The choice of action resolution is an important hyperparameter. For the LIBERO benchmark, we find that a resolution of 1000 is sufficient. Decreasing the resolution to 250 degrades performance, reducing the success rate by 1.5 points, while a higher resolution of 4000 yields no additional performance gains.

\paragraph{Image Tiling}
When providing multiple image observations to the VLM, one can either tile them into a single composite image or feed them as separate inputs. We find that this decision has no discernible impact on performance.
\input{tables/ablation}

%% file: tables/ablation.tex
\begin{table}[!t]
 \centering
 \setlength\tabcolsep{4pt}
 \setlength{\abovetopsep}{5pt}
 \begin{tabular}{ccccccc}
 \toprule
  \thead{Row\\ID} & \thead{Ensemble\\Act.} & \thead{Masked \\ Act. Aug.} & \thead{Tiled\\Img.} & \thead{Act.\\Res.} & \thead{Avg.\\ Succ.} & \thead{$\Delta$ perf.} \\
 \midrule
 0 & \yes & \yes & \yes & 1000 & 94.7 & 0.0 \\
 1 & \no  & \yes & \yes & 1000 & 92.0 & -2.0 \\
 2 & \yes & \no  & \yes & 1000 & 93.5 & -1.2 \\
 3 & \yes & \yes & \yes & 4000 & 94.2 & -0.5 \\
 4 & \yes & \yes & \yes & 250  & 93.2 & -1.5 \\
 5 & \yes & \yes & \no  & 1000 & 94.5 & -0.2 \\
 \bottomrule
 \end{tabular}
 \caption{\tb{Ablations.} We ablate various design choices for \method on LIBERO. We report the average success rate over the four LIBERO suites.}
 \label{table:ablation}
\end{table}

%% file: sec/5_conclusions.tex
\vspace{-2mm}
\section{Conclusions and Limitations}
\label{sec:conclusion}
In this work, we made the case for a simple VLA design that preserves the integrity of the base VLM without altering its tokenization or introducing new architectural components. We demonstrated that with the right recipe, this approach outperforms more involved strategies—a surprising result that considering prevailing trends in the literature. 

Despite these promising findings, our work has limitations that present clear directions for future research. A key area to explore is how \method would perform when trained with large-scale action data. Another area of investigation would be to improve inference speed of \method using optimization techniques like quantization and distillation.